\documentclass[letterpaper]{article} 

\usepackage{aaai24}  
\usepackage{amsmath} 
\usepackage{times}  
\usepackage{helvet}  
\usepackage{courier}  
\usepackage[hyphens]{url}  
\usepackage{graphicx} 
\usepackage{natbib}  
\usepackage{caption} 
\usepackage{algorithm}
\usepackage{algorithmic}

\usepackage{newfloat}
\usepackage{listings}
\usepackage{xcolor}

\DeclareCaptionStyle{ruled}{labelfont=normalfont,labelsep=colon,strut=off} 
\floatstyle{ruled}
\newfloat{listing}{tb}{lst}{}
\floatname{listing}{Listing}

\lstdefinestyle{default_style}{
	basicstyle={\footnotesize\ttfamily},
	numbers=left,numberstyle=\footnotesize,xleftmargin=2em,
	aboveskip=0pt,belowskip=0pt,%
	showstringspaces=false,tabsize=2,breaklines=true,
    frame=false,
    backgroundcolor=\color{white}
}

\lstdefinestyle{prompt_style}{
  backgroundcolor=\color{lightgray}, 
  basicstyle=\ttfamily\footnotesize, 
  keywordstyle=\color{blue}, 
  commentstyle=\color{green}, 
  stringstyle=\color{red}, 
  frame=single, 
  breaklines=true, 
  captionpos=b, 
}

\title{LegalGuardian: A Privacy-Preserving Framework for Secure \\Integration of Large Language Models in Legal Practice}
\author{
    M. Mikail Demir,
    Hakan T. Otal,
    M. Abdullah Canbaz
}
\affiliations{
    University at Albany, SUNY\\
    Department of Information Science and Technology\\
    Albany, New York, 12222, USA\\
    mdemir@albany.edu, hotal@albany.edu, mcanbaz@albany.edu
}

\begin{document}

\maketitle

\begin{abstract}
Large Language Models (LLMs) hold promise for advancing legal practice by automating complex tasks and improving access to justice. However, their adoption is limited by concerns over client confidentiality, especially when lawyers include sensitive Personally Identifiable Information (PII) in prompts, risking unauthorized data exposure. To mitigate this, we introduce LegalGuardian, a lightweight, privacy-preserving framework tailored for lawyers using LLM-based tools. LegalGuardian employs Named Entity Recognition (NER) techniques and local LLMs to mask and unmask confidential PII within prompts, safeguarding sensitive data before any external interaction. We detail its development and assess its effectiveness using a synthetic prompt library in immigration law scenarios. Comparing traditional NER models with one-shot prompted local LLM, we find that LegalGuardian achieves a F1-score of 93\% with GLiNER and 97\% with Qwen2.5-14B in PII detection. Semantic similarity analysis confirms that the framework maintains high fidelity in outputs, ensuring robust utility of LLM-based tools. Our findings indicate that legal professionals can harness advanced AI technologies without compromising client confidentiality or the quality of legal documents.
\end{abstract}

\section{Introduction}

Transformer-powered LLMs are fundamentally transforming conversational AI by offering unparalleled fluency and deep contextual understanding~\citep{taecharungroj_what_2023}. In the traditionally conservative and complex legal sector, LLMs are pioneering change by automating intricate tasks such as predicting legal judgments~\citep{medvedeva-mcbride-2023-legal,chalkidis-etal-2019-neural}, analyzing vast legal documents~\citep{trautmann2023largelanguagemodelprompt,mamakas2022processinglonglegaldocuments}, and generating sophisticated legal writings~\citep{guha2023legalbenchcollaborativelybuiltbenchmark}. This technological revolution holds immense promise for democratizing legal services. By breaking down barriers of income, language, and geography, LLMs could play a pivotal role in addressing the global access-to-justice crisis, where countless individuals lack adequate legal support~\citep{chien2024generative}.

Despite the transformative potential of LLM-based chatbots in the legal sector, their widespread adoption faces significant barriers, foremost among them the risk of compromising client confidentiality. Confidentiality is a cornerstone of legal ethics, enshrined in professional obligations~\citep{Fischel1998LawyersAC}. The American Bar Association (ABA) Model Rules of Professional Conduct~\citep{ABA_Model_Rules}, which form the foundation of nearly all state rules, include Rule 1.6, which prohibits the disclosure of information related to the representation of a client without the client’s consent. Moreover, Comment 18 elaborates that lawyers must act competently to safeguard client information against inadvertent or unauthorized disclosure, whether by the lawyer, others involved in the representation, or individuals under the lawyer’s supervision~\citep{ABA_Model_Rules}. Furthermore, Comment 19 advises attorneys to take reasonable precautions to prevent confidential information from being accessed by unintended recipients during communications~\citep{ABA_Model_Rules}.

While these confidentiality obligations are universally accepted for a long period of time, they have evolved with each new technological advancement, bringing fresh concerns. For example in 2010s, with the emergence of internet-based legal research,~\citet{klinefelter_when_nodate}, examined online tracking of search queries in the context of confidentiality risks for lawyers. In this regard, in-context learning through natural language prompting has introduced a new paradigm, enabling professionals to perform tasks such as data annotation, search, and question-answering, while giving rise to new concerns for confidentiality and data security~\citep{yu-etal-2023-exploring}. However, lawyers often include PII or sensitive client data in their prompts by incorporating excerpts from case decisions, contracts, or client correspondence. For instance, an attorney working on an immigration case might include details such as a client’s name, nationality, immigration status, or specific travel history to enhance the chatbot’s response accuracy.

Including sensitive information in prompts risks unauthorized exposure to third-party LLM providers, potentially breaching attorney-client privilege and data protection laws. Recent precedents highlight these concerns. A recent incident on March 24, 2023, highlighted the vulnerabilities associated with LLMs in legal contexts. Due to weaknesses in the Redis client open-source library~\citep{ChatgptNews} , OpenAI’s ChatGPT~\citep{openai2024gpt4technicalreport} inadvertently exposed users’ chat histories.

Such instances of data leaks involving LLMs have raised significant concerns in both professional and public domains. To address how existing professional rules apply in the context of generative AI, ABA published Formal Opinion 512 on July 29, 2024~\citep{formalopinion512}, on the use of generative AI. The opinion warns that self-learning generative AI tools inherently pose risks to confidentiality, noting that information inputted about one client may later surface in responses to unrelated prompts, potentially exposing sensitive data to other users, clients, courts, or third parties.

Several states—including California, Florida, Illinois, Kentucky, Maryland, Minnesota, New York, Texas, and Virginia— followed ABA's stance and have issued guidance or established task forces to tackle the ethical implications of AI in legal practice~\citep{stateaiforce}. While these initiatives are commendable, they collectively underline a critical concern: existing technologies do not yet offer sufficient safeguards to meet the rigorous confidentiality standards required in the legal profession.

While many large law firms possess the resources to develop and deploy proprietary LLM models tailored to their specific needs \citep{dentonsDentonsLaunch}, the majority of legal professionals—ranging from legal aid workers to solo practitioners—lack access to such advanced and costly infrastructure. This disparity underscores the need for lightweight, accessible frameworks that enable these practitioners to harness the potential of LLMs while adhering to strict confidentiality requirements and ethical obligations.

To address these challenges, we propose LegalGuardian, a lightweight privacy-preserving framework tailored for the lawyer-side use of LLM-based chatbots. The framework leverages Named Entity Recognition (NER) techniques and local LLMs to mask and unmask confidential PII within prompts, ensuring client confidentiality before any interaction with external systems. This streamlined, rule-based approach allows legal professionals to utilize LLM tools while maintaining their ethical duty of competence.

Our experimental setup employs a set of synthetic prompts we have generated for this task, that simulates workflow of an immigration lawyer, covering scenarios such as visa applications, asylum cases, and naturalization processes. By combining advanced NER techniques with local LLM-prompted PII detection, we evaluated the framework across three metrics—accuracy, entity-level precision/recall, and semantic similarity—to assess its privacy-utility trade-off. Detailed results are presented in the subsequent sections.

\subsection{Related Work}

The integration of LLMs into legal practice intersects disciplines like computer science, computational linguistics, cryptography, privacy, and ethics. Preserving client confidentiality while leveraging LLMs requires understanding these fields. Researchers have explored techniques such as differential privacy, data sanitization, encryption methods, and federated learning to mitigate LLM privacy risks~\citep{edemacu_privacy_2024}. In the legal domain, strict ethical obligations and regulations amplify these challenges, necessitating innovative solutions. This section presents related works most relevant to our study.

\textbf{Cryptographic Techniques:} Cryptographic methods like Multi-Party Computation (MPC)~\citep{4568388} and Homomorphic Encryption (HE)~\citep{cryptoeprint:2012/099} secure computations on sensitive data. MPC allows collaborative computation without sharing inputs~\citep{Lindell2020SecureMC}, and HE enables computations on encrypted data without decryption~\citep{Iezzi2020PracticalPD}. However, these techniques often have high computational overhead and scalability issues, especially with high-dimensional free-text legal data. Latency from cryptographic operations can make real-time applications impractical~\citep{Yavuz2017RealTimeDS}, and implementing them with LLMs requires substantial architectural changes, challenging practical deployment.

\textbf{Differential Privacy and Adversarial Training:} Differential Privacy (DP)~\citep{dwork_differential_2006} controls privacy loss by adding calibrated noise to data or query results, limiting the impact of any single individual's data~\citep{Kifer2011NoFL}. Applying DP to free-text data is challenging due to natural language complexity~\citep{Li2021LargeLM}. Adversarial training~\citep{li2023mpcformerfastperformantprivate,coavoux2018privacypreservingneuralrepresentationstext} learns representations predictive for the main task but invariant to private attributes, aiming to prevent sensitive information leakage. However, these methods often trade off utility for privacy, degrading main task performance, especially when private information is intertwined with task-relevant features~\citep{zhou-etal-2022-textfusion}.

\textbf{Encryption-Based Methods:} Techniques like EmojiCrypt~\citep{lin2024emojicryptpromptencryptionsecure} protect user inputs by substituting sensitive text with encrypted representations (e.g., emojis), obscuring PII while preserving structure for LLM processing. However, these methods risk the LLM misinterpreting encrypted tokens, reducing performance or causing unintended outputs~\citep{edemacu_privacy_2024}. Non-standard tokens can introduce ambiguity, affecting model understanding, especially in precision-critical legal contexts.

\textbf{Federated Learning:} Federated Learning (FL)~\citep{DBLP:journals/corr/McMahanMRA16} trains models across decentralized devices holding local data, keeping sensitive data on local devices—a benefit in privacy-sensitive domains like legal practice~\citep{chalamala_federated_2022}. In LLMs, FL could allow legal organizations to train models without exposing proprietary data. However, FL addresses privacy during training, not inference, where privacy breaches often occur with LLM chatbots. FL also introduces significant communication overhead and requires node synchronization, impractical in heterogeneous legal environments~\citep{kairouz2021advancesopenproblemsfederated}. Modifying LLM architectures for FL further limits its applicability.

\textbf{Privacy in Legal Applications of LLMs:} Applying LLMs in legal contexts faces challenges due to strict confidentiality and sensitive data. Research has adapted privacy-preserving techniques for legal needs, such as combining legal ontologies with NER to enhance sensitive information redaction~\citep{Cardellino2017ALH}. Domain-specific models like Legal-BERT~\citep{chalkidis-etal-2020-legal} improve understanding of legal terminology, aiding PII detection. However, these methods struggle with scalability and handling nuanced legal language. There is a lack of comprehensive solutions balancing privacy, efficiency, and usability tailored for legal practitioners using LLMs.

\textbf{Data Sanitization:} Data sanitization identifies and removes PII from user input before processing. \citet{kan2023protectinguserprivacyremote} proposed the PP-TS framework, using a local LLM to detect and mask sensitive attributes before sending data to a cloud-based LLM, involving pre-processing, LLM invocation, and post-processing. Similarly, \citet{chen2023hideseekhaslightweight} introduced the "Hide and Seek" (HaS) framework, requiring two models—a masker for anonymization and a reconstructor for de-anonymization—to protect privacy while maintaining utility. While effective in some contexts, these methods have limitations: training multiple models increases computational costs and complexity, making them less feasible for many legal practices. Additionally, PII detection accuracy and potential loss of contextual information can impact downstream task performance~\cite{zhou-etal-2022-textfusion}.

\subsection{Introduction to LegalGuardian Framework} 
\begin{figure}[!b]
    \centering
    \includegraphics[width=\linewidth]{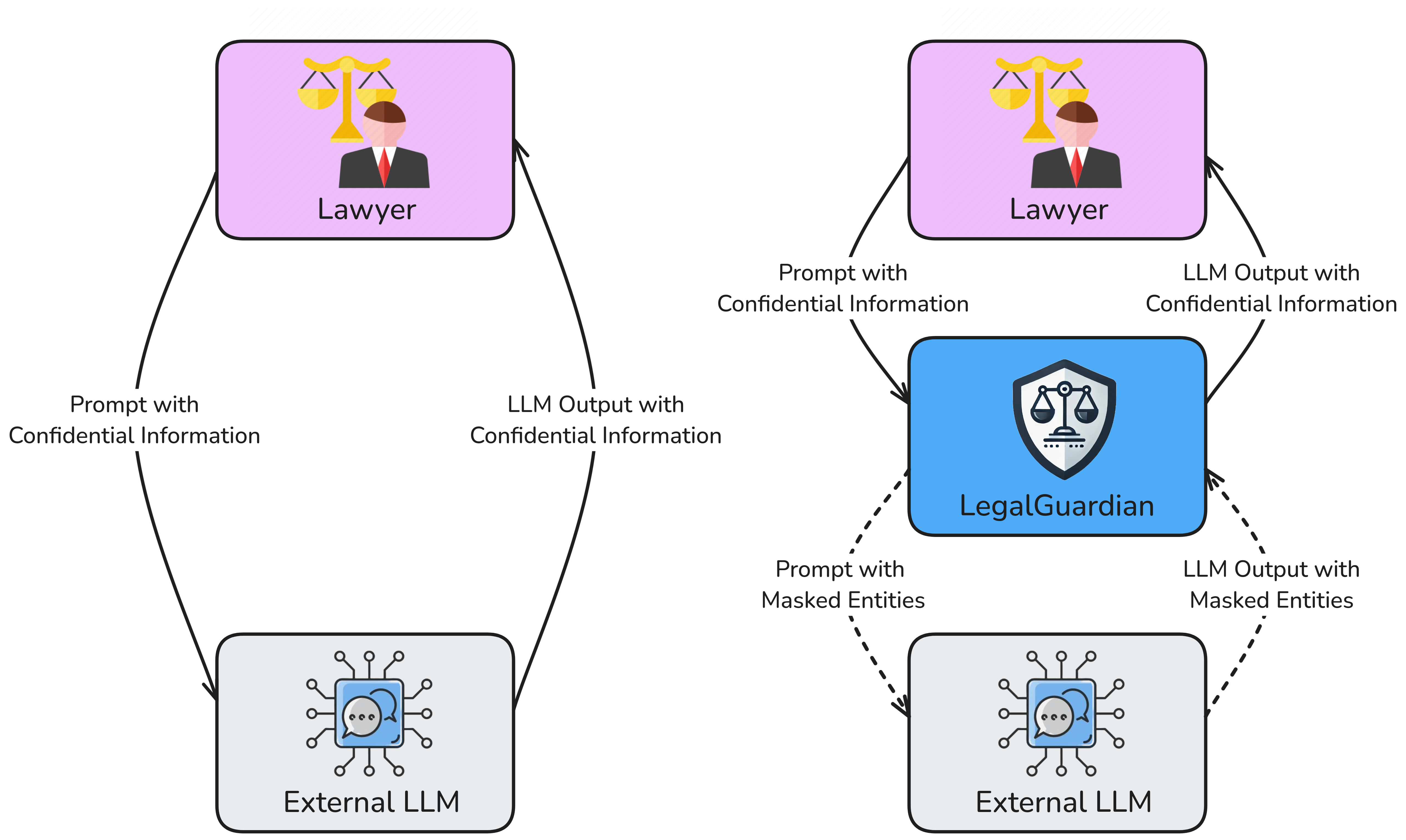}
    \caption{Comparison of Legal Prompting Without (Left) and With (Right) the LegalGuardian Framework}
    \label{fig:legalguardian}
\end{figure}

\begin{figure*}[!htb]
    \centering
    \includegraphics[width=\linewidth]{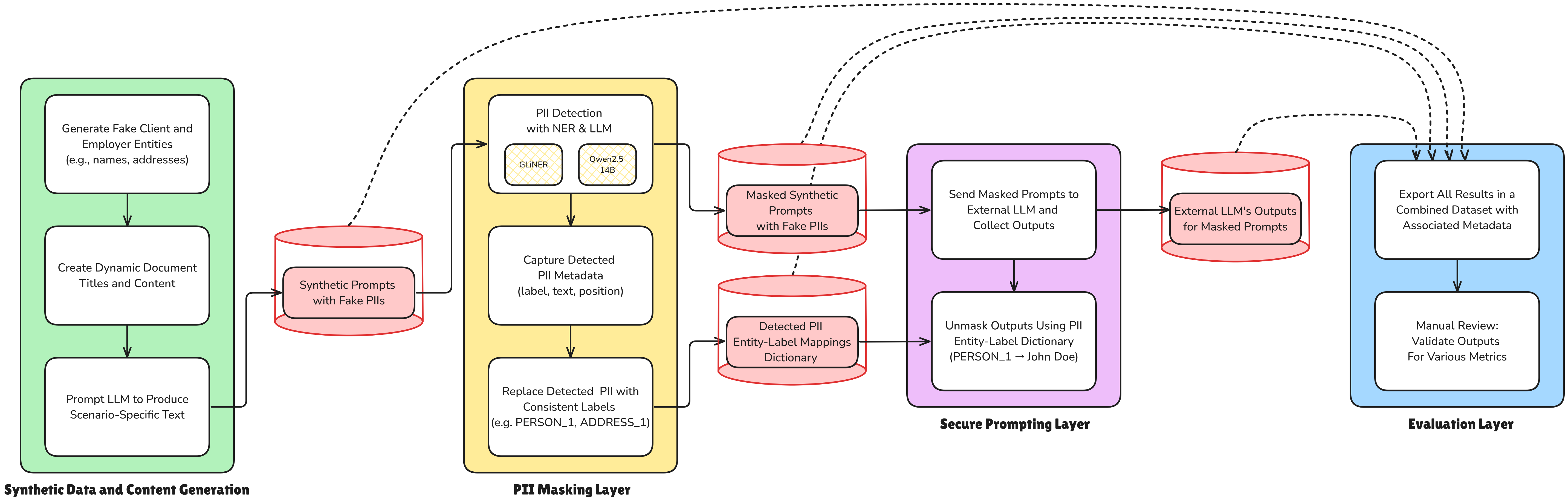}
    \caption{Dataset Creation and Evaluation Pipeline for LegalGuardian Framework}
    \label{fig:pipeline-diagram}
\end{figure*}
Maintaining client confidentiality is a fundamental ethical obligation in the legal profession. To facilitate the integration of AI-enabled systems without compromising this confidentiality, we present \textit{LegalGuardian}, a novel privacy-preserving framework specifically designed for legal practitioners. In this section, we provide a comprehensive overview of LegalGuardian’s development and evaluation, including a detailed case study on privacy-preserving techniques, the framework’s architectural design, the creation of a synthetic legal prompt dataset, and an assessment of its effectiveness in protecting attorney-client confidentiality.

To this end, first, we conduct a comprehensive case study on the application of privacy-preserving techniques within the legal domain, specifically addressing the protection of attorney-client confidentiality and privacy. This study aims to pave the way for the broader adoption of generative AI tools in the legal sector by tackling critical privacy concerns that hinder their integration.

Second, we develop a synthetic legal prompt dataset consisting of 50 synthetically generated prompts. These prompts are tailored to reflect realistic scenarios faced by immigration lawyers practicing in the United States. We aim to share this dataset with the research community to support further studies in this domain.

Third, we introduce LegalGuardian, displayed in Figure~\ref{fig:legalguardian}, the first privacy protection framework explicitly designed to safeguard attorney-client confidentiality in the era of generative AI. LegalGuardian employs NLP techniques to mask and unmask confidential PII within prompts, ensuring sensitive data remains secure during interactions with Large Language Model (LLM)-based chatbots.

Finally, we provide a thorough assessment of the proposed technique’s effectiveness in protecting confidentiality from both technical and legal perspectives. Technically, we evaluate the framework using metrics such as accuracy, entity-level accuracy, and contextual similarity. From a legal standpoint, we examine how our approach aligns with existing laws and ethical guidelines governing attorney-client privilege and confidentiality.

\section{Methodology}

In this section, we outline the methodology employed to develop and evaluate LegalGuardian, our proposed privacy-preserving framework designed for masking and unmasking PII in legal prompts. Our approach integrates the generation of synthetic data representative of real-world legal scenarios, the application of advanced NER techniques, and the utilization of local LLMs for PII detection through one-shot prompting. We also establish a comprehensive evaluation protocol to ensure that our framework effectively preserves privacy while maintaining utility in legal NLP tasks.

\subsection{Overview of LegalGuardian Framework}

To effectively safeguard client confidentiality while utilizing LLMs in legal practice, we developed the LegalGuardian framework. Figure~\ref{fig:pipeline-diagram} illustrates the data creation and evaluation pipeline of our framework, which consists of four main components:

\begin{enumerate}
\item \textbf{Synthetic Data and Content Generation (Green Box)}: We generate realistic legal prompts that simulate various immigration-related scenarios, providing a comprehensive dataset for testing and evaluation.
\item \textbf{PII Masking Layer (Yellow Box)}: Using advanced NER models and local LLMs, we identify and mask PII entities within the prompts to prevent the disclosure of sensitive information.
\item \textbf{Secure Prompting Layer (Purple Box)}: With the PII securely masked, we interact with external LLMs using these sanitized prompts to perform downstream NLP tasks without risking privacy breaches.
\item \textbf{Evaluation Layer (Blue Box)}: We assess the effectiveness of the framework by evaluating its ability to preserve privacy while maintaining the functional utility of the outputs.
\end{enumerate}

\subsection{Synthetic Data and Content Generation (Green Box)}

To simulate realistic legal scenarios for testing our framework, we generated a synthetic dataset specifically designed to reflect the workflow of an immigration lawyer practicing in the United States.

Initially, we synthesized realistic client and employer details, including names, addresses, nationalities, and other pertinent metadata such as visa types and practice areas. The \texttt{Faker}~\cite{fakerWelcomeFakerx2019s} library was employed to generate these details, ensuring diversity and authenticity within the dataset. Random selection methods were applied to probabilistically assign document titles and subfields within practice areas, thereby enhancing the variability and representativeness of the prompts.

Subsequently, dynamic prompts were constructed by integrating the generated entities and metadata into structured templates. These prompts were then processed using the \texttt{Qwen-2.5 14B } language model ~\cite{qwen2.5}, producing 50 realistic prompts that reflect various immigration case scenarios, such as visa applications, asylum cases, and naturalization processes. Detailed prompt templates used in this process are provided in Appendix.

We selected 50 prompts for the study to balance the need for a representative dataset with the practical constraints of conducting detailed manual reviews for each prompt. This approach enabled the creation of a diverse and realistic dataset for evaluating the privacy-preserving capabilities of the framework.

\begin{table}[!t]
\begin{lstlisting}[label=pro:prompt1, caption=System prompt given to LLM for doing PII entity recognition from given text]
You are a Named Entity Recognition bot. 
Given a paragraph, you identify entity labels: ["person", "case_number", "date_of_birth", "address", "company", "tax ID", "location", "date", "law office", "nationality"]
Dont provide any explanations or comments, only use the given text to detect entities.

Example: For given text "My name is John Doe, I live in London.", output should be:
{
    "entities": [
        "John Doe": "person",
        "London": "location"
    ]
}

The output must be strictly in JSON format as follows:
{{
    "entities": [
        {"<entity_name>": "<entity_label>"},
        {"<entity_name>": "<entity_label>"},
        ...
    ]
}}
\end{lstlisting}
\vspace{-2mm}
\end{table}

\subsection{PII Recognition \& Masking Layer (Yellow Box)}

To ensure the confidentiality of sensitive information within the synthetic prompts, we implemented a PII masking layer comprising two separate components to compare their results at the end. 

Firstly, we utilized GLiNER \citep{zaratiana2023glinergeneralistmodelnamed}, specifically GLiNER\_Multi\_PII-v1 model, which is a NER model designed for flexibility and efficiency, capable of identifying custom entity types by leveraging BERT instead of being limited to predefined entity categories like traditional NER models. This choice was motivated by our task’s focus on PII detection, aligning perfectly with the model’s fine-tuned capabilities for recognizing and handling sensitive personal information. We chose GLiNER because our domain-specific task required assigning unique, custom entity names rather than relying on generic categories provided by other models, enabling more precise handling of legal-specific PII.

Secondly, we employed the \texttt{Qwen2.5-14B} language model~\cite{qwen2.5}, prompting it to identify PII entities in a one-shot manner, as illustrated in Listing~\ref{pro:prompt1}. Qwen-2.5 14B model is chosen because it is a state-of-the-art LLM that consistently ranks at the top of LLM leaderboards for its size category, making it an ideal balance between performance and computational efficiency. Among medium-sized open-source models (14B parameters), Qwen-2.5 demonstrated the best overall performance across benchmarks, solidifying it as the optimal choice for our application.

\subsubsection{The Masking Process:}

\begin{enumerate}
    \item \textbf{PII Detection}: We applied both the GLiNER and the Qwen2.5-14B language model to the synthetic prompts to detect PII entities. For each result, we recorded each detected entity along with its text, label, and position within the prompt. Then, a dictionary is used to store the original entity with its respective label, for both model.
    \item \textbf{Entity Replacement}: Detected PII entities were systematically replaced with consistent placeholder labels (e.g., \texttt{[PERSON\_1]}, \texttt{[ADDRESS\_1]}) to ensure anonymization while maintaining the coherence of the text. This process involved assigning unique labels to each distinct entity within the dataset. A custom Python function, supported by a dictionary that stored original entities along with their pseudonymized labels, ensured that the same entity was consistently replaced with the same placeholder throughout the prompt. By combining the entity type with a sequential number (e.g., \texttt{[PERSON\_1]}, \texttt{[PERSON\_2]}), this approach preserved the relationships and coherence between entities in the text while enabling accurate masking.
    \item \textbf{Entity Dictionary Maintenance}: We maintained two entity dictionaries for each model, mapping the original entities to their corresponding placeholders. This allowed for potential re-identification (unmasking) in later stages if necessary, enabling downstream tasks that might require access to the original data under controlled conditions.
\end{enumerate}

By effectively masking the PII at this stage, we ensured that sensitive data was protected before any interaction with external systems or models. This approach maintains confidentiality and complies with legal and ethical standards for data protection. The performance of the sensitive data masking process was evaluated at the end of the pipeline to assess its effectiveness.

\subsection{Secure Prompting Layer (Purple Box)}

After effectively masking the PII in the prompts, we implemented a secure prompting mechanism—referred to as the Secure Prompting Layer (illustrated in the Purple Box of Figure~\ref{fig:pipeline-diagram})—to interact with an external LLM. This layer enables the use of masked prompts to perform downstream NLP tasks such as summarization, translation, or legal analysis without risking the exposure of sensitive information.

\subsubsection{Secure Prompting Process:}

\begin{enumerate}
\item \textbf{Model Interaction}: The masked prompts are transmitted to the external LLM for processing. Since all PII has been replaced with placeholders, this interaction adheres to the established privacy constraints, ensuring that no confidential data is disclosed during communication. For the purposes of simulation and evaluation, we utilized a local instance of the Qwen2.5-14B LLM, treating it as an external LLM to mimic real-world conditions.
\item \textbf{Post-processing of Outputs}: Upon receiving the outputs generated by the LLM, we perform post-processing to unmask PII where necessary. This involves replacing the placeholders in the LLM outputs with the original entities using the maintained entity dictionaries. For example, a placeholder like \texttt{[PERSON\_1]} is substituted with the actual name "John Doe." This step restores the contextual integrity of the output while ensuring that sensitive information was protected during the interaction with the external LLM.
\end{enumerate}

By employing this approach, we leverage the advanced capabilities of the LLM for various NLP tasks while maintaining strict confidentiality of sensitive information throughout the entire process. This method ensures compliance with legal and ethical standards, allowing legal professionals to benefit from AI advancements without compromising client privacy.

\subsection{Evaluation Layer (Blue Box)}

To rigorously assess the effectiveness of the LegalGuardian framework, we implemented a comprehensive evaluation layer focusing on several key metrics designed to measure both privacy preservation and utility:

\begin{enumerate}
    \item \textbf{Masking Accuracy}: We evaluated the precision and recall of the PII masking process to determine how accurately the framework identifies and masks sensitive entities.
    \item \textbf{Semantic Consistency}: We measured the semantic similarity between the outputs generated by the LLM when using the original (unmasked) prompts and the masked prompts. This assessment ensured that the masking process did not adversely affect the semantic content and overall utility of the outputs.
\end{enumerate}

All data from the prompts, entity dictionaries, and outputs were systematically stored and organized using the \texttt{Pandas} library, and metrics were calculated using \texttt{SpaCy}~\cite{spacy}, facilitating thorough analysis. Both automated metrics and manual reviews were employed to validate the framework’s effectiveness in preserving privacy while maintaining functional utility in legal NLP tasks.


\section{Results \& Findings}
To evaluate the experimental performance of Qwen and GLiNER in terms of their accuracy in masking and unmasking, as well as to examine the critical balance between preserving privacy and maintaining utility, we conducted a comprehensive analysis based on three key metrics:
\begin{enumerate}
    \item \textbf{Overall Accuracy}: This metric includes precision, recall, and F1 scores, providing quantitative insights into each model’s ability to accurately detect and mask sensitive entities across the dataset.
    \item \textbf{Entity-Level Accuracy}: Focusing on individual entity categories, this metric assesses the detection and masking performance for specific types of sensitive information. By offering a granular view of each model’s behavior with respect to entities such as \textit{person}, \textit{case\_number}, and \textit{address}, we can identify strengths and weaknesses in their handling of different PII categories.
    \item \textbf{Semantic Similarity}: This metric evaluates the extent to which the masking process affects the semantic integrity of the outputs. By comparing the results obtained from masked prompts with those from unmasked prompts, we quantify any loss of semantic content or degradation in output fidelity introduced by the masking procedure.
\end{enumerate}
\subsection{Overall Accuracy}

The masking and unmasking accuracies of the Qwen and GLiNER models were evaluated using a manually curated dataset comprising 50 prompts for each model. For every prompt, detailed review tables were generated, documenting the detected entity categories, the expected categories, and the correctness of entity detection.

The evaluation focused on specific entity categories relevant to legal applications, including \textit{person}, \textit{case\_number}, \textit{date\_of\_birth}, \textit{address}, \textit{company}, \textit{tax ID}, \textit{location}, \textit{date}, \textit{law office}, and \textit{nationality}. True positives (TP), false positives (FP), and false negatives (FN) were identified to calculate precision, recall, and F1 scores. Here, true positives represented correctly detected entities; false positives indicated entities incorrectly detected or over-detected; and false negatives corresponded to entities that were expected but missed by the models. In total, the dataset includes different 460 entities.

The overall results demonstrated that GLiNER achieved higher precision due to fewer false positives, highlighting its effectiveness in avoiding over-detection. In contrast, Qwen exhibited slightly better recall by successfully identifying more relevant entities, albeit with a higher rate of over-detection. The F1 scores reflected a balanced trade-off between precision and recall for both models.

\begin{figure}[!t]
    \centering
    \includegraphics[width=\linewidth]{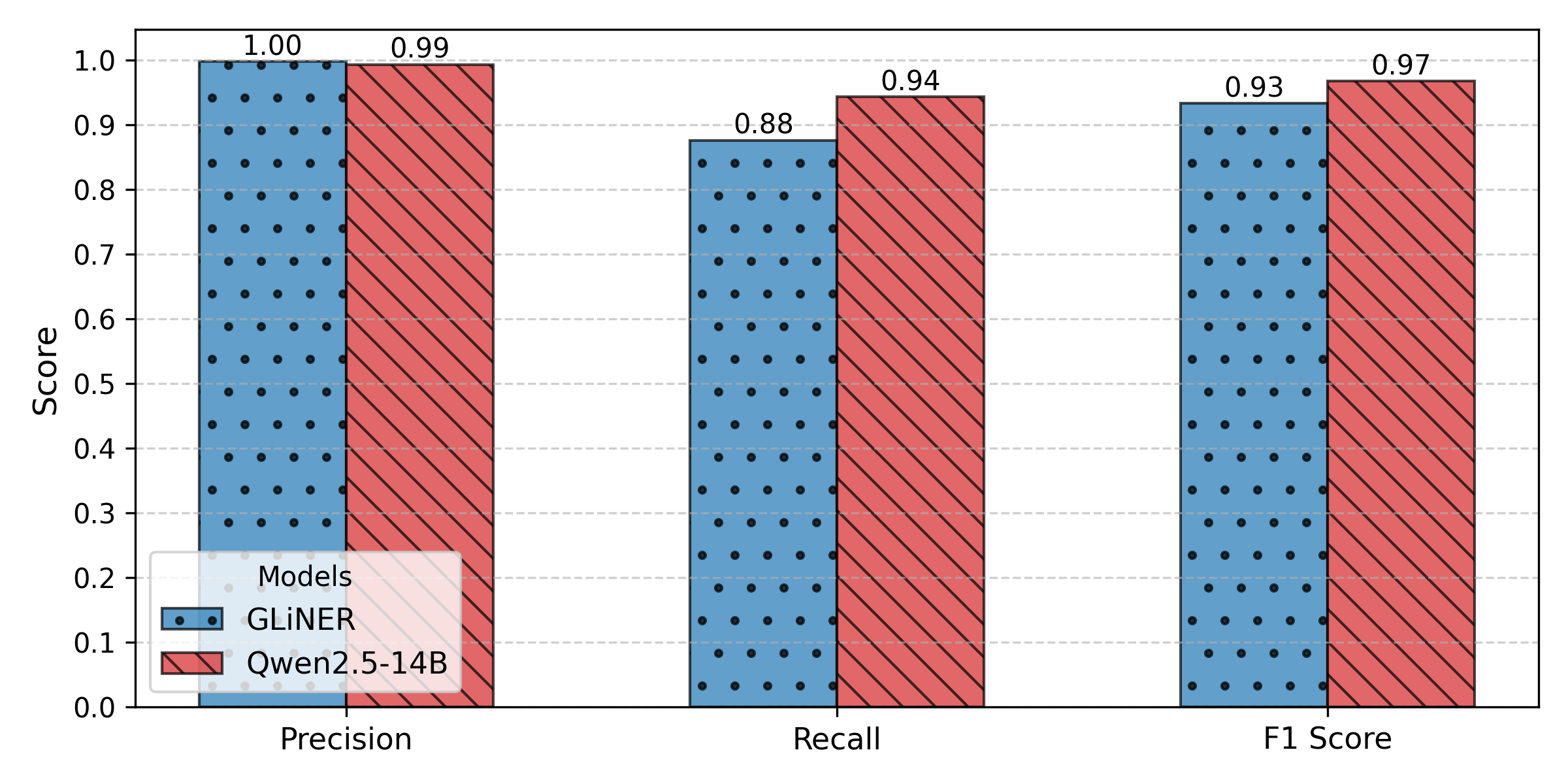}
    \caption{Overall performance metrics for GLiNER and Qwen2.5-14B}
    \label{fig:model-metrics}
    \vspace{-2mm}
\end{figure}

Figure ~\ref{fig:model-metrics} compares the performance of two models, GLiNER and Qwen2.5-14B, based on three metrics: Precision, Recall, and F1 Score. GLiNER achieves a perfect Precision of 100\% compared to 99\% for Qwen2.5-14B, highlighting its stronger ability to avoid false positives. However, Qwen2.5-14B outperforms GLiNER in Recall (94\% vs. 88\%), indicating better identification of relevant instances. Similarly, Qwen2.5-14B exhibits a higher F1 Score (97\% vs. 93\%), reflecting its superior overall balance between precision and recall. This suggests that while GLiNER is more precise, Qwen2.5-14B demonstrates better all-around performance.

\subsection{Entity-Level Accuracy Metrics}

To gain a deeper understanding of the models’ performance across specific entity categories, we computed entity-level precision, recall, and F1 scores for both Qwen2.5-14B and GLiNER. This granular analysis was crucial to evaluate how effectively each model handled different types of sensitive information commonly encountered in legal contexts, such as \textit{person}, \textit{address}, \textit{date\_of\_birth}, \textit{law\_office}, and other relevant categories. By examining these metrics, we identified variations in model behavior across entity types, revealing strengths and weaknesses in detecting particular categories.

Table~\ref{tab:comparison} shows the detailed performance metrics for GLiNER and Qwen2.5-14B across various entity types, highlighting key distinctions in their strengths. GLiNER consistently excels in precision, particularly for structured entities like \textit{case\_number} and \textit{tax\_id}, achieving perfect scores in precision, recall, and F1. This underscores its capability to minimize false positives, making it highly effective for well-defined categories. Similarly, GLiNER maintained perfect precision for entities such as \textit{address} and \textit{company}, further demonstrating its reliability in handling structured information.

Qwen2.5-14B, on the other hand, exhibited higher recall for more context-dependent categories like \textit{person} and \textit{address}. This suggests its strength in identifying a broader range of relevant entities, even at the cost of a slight reduction in precision. For instance, in the \textit{person} category, Qwen2.5-14B achieved a recall of 77\% compared to GLiNER's 72\%, while maintaining equal precision of 100\%. This trade-off highlights Qwen2.5-14B's proficiency in broader entity detection, particularly for less structured contexts.

In table~\ref{tab:comparison}, both models show balanced performance for core entities such as \textit{date} and \textit{location}. GLiNER achieved a perfect F1 score for \textit{date}, while Qwen2.5-14B delivered a perfect F1 score for \textit{location}, reflecting their shared reliability for these crucial categories. Additionally, for entities such as \textit{law\_office} and \textit{case\_number}, both models attained perfect scores across all metrics, confirming their strong capabilities for well-defined, structured types.

Overall, we observe complementary strengths between the two models. While GLiNER’s precision-driven approach is advantageous for structured entity types, Qwen2.5-14B’s higher recall makes it better suited for detecting context-dependent or ambiguous entities.

\begin{table}[!t]
    \centering
    \renewcommand{\arraystretch}{1.2}
    \begin{tabular}{|l|c|c|c|c|c|c|}
        \hline
        \textbf{Entity Type} & \multicolumn{3}{c|}{\textbf{GLiNER}} & \multicolumn{3}{c|}{\textbf{Qwen2.5-14B}} \\
        \cline{2-7}
         & \textbf{P} & \textbf{R} & \textbf{F1} & \textbf{P} & \textbf{R} & \textbf{F1} \\
        \hline
        person & 1.00 & 0.72 & 0.84 & 1.00 & 0.77 & 0.87 \\
        \hline
        address & 1.00 & 0.98 & 0.99 & 0.99 & 1.00 & 0.99 \\
        \hline
        nationality & 0.98 & 0.91 & 0.95 & 0.98 & 1.00 & 0.99 \\
        \hline
        date & 1.00 & 1.00 & 1.00 & 0.98 & 0.89 & 0.93 \\
        \hline
        location & 1.00 & 0.65 & 0.79 & 1.00 & 1.00 & 1.00 \\
        \hline
        law\_office & 1.00 & 1.00 & 1.00 & 1.00 & 1.00 & 1.00 \\
        \hline
        company & 1.00 & 0.91 & 0.95 & 1.00 & 1.00 & 1.00 \\
        \hline
        tax\_id & 1.00 & 1.00 & 1.00 & 1.00 & 1.00 & 1.00 \\
        \hline
        case\_number & 1.00 & 1.00 & 1.00 & 1.00 & 1.00 & 1.00 \\
        \hline
    \end{tabular}
    \caption{Comparison of GLiNER and Qwen2.5-14B metrics across all entity types}
    \label{tab:comparison}
    \vspace{-2mm}
\end{table}

\subsection{Semantic Similarity}

To assess the impact of the masking and unmasking processes on the semantic integrity of the outputs, we conducted a semantic similarity analysis. This analysis evaluated how closely the unmasked outputs from the LegalGuardian pipeline aligned with the original outputs generated from unmasked prompts. The objective was to ensure that masking sensitive information did not compromise the quality or meaning of the outputs—a critical factor in maintaining utility while preserving privacy.

We employed three measures to compute semantic similarity:
\begin{itemize}
    \item \textbf{Cosine Similarity}:  Implemented using the \texttt{spacy-transformer-md} model, this metric measures the similarity between sentence embeddings of the original and unmasked outputs. It captures the overall semantic alignment by comparing the vector representations of the texts.
    \item \textbf{Jaro-Winkler Distance}: This string-based metric focuses on character-level similarities, making it particularly useful for detecting slight variations in the output text. It assesses how similar two strings are by considering the number and order of matching characters.
    \item \textbf{Levenshtein Distance}: Also a character-level metric, the Levenshtein distance quantifies the minimum number of single-character edits—insertions, deletions, or substitutions—required to change one string into the other. It provides insights into the structural changes introduced by the masking and unmasking processes.
\end{itemize}

The evaluation process involved generating baseline outputs by inputting the original, unmasked prompts into the Qwen model. These baseline outputs were then compared with the unmasked outputs obtained from each model (GLiNER and Qwen2.5-14B) after processing through the LegalGuardian pipeline. By comparing these outputs, we measured the extent to which the pipeline preserved the semantic meaning of the text.

\begin{table}[!t]
    \centering
    \renewcommand{\arraystretch}{1.2}
    \begin{tabular}{|l|c|c|}
        \hline
        \textbf{Metrics} & \multicolumn{2}{c|}{\textbf{Mean Score}} \\
        \cline{2-3}
         & \textbf{GLiNER} & \textbf{Qwen2.5-14B} \\
        \hline
        Cosine Similarity & \textbf{0.9808} & 0.9731 \\
        \hline
        Jaro-Winkler Similarity & \textbf{0.8328} & 0.7601 \\
        \hline
        Levenshtein Distance & 0.4358 & \textbf{0.4017} \\
        \hline
    \end{tabular}
    \caption{Similarity and distance metrics for measuring Semantic Consistency}
    \label{tab:metrics}
    \vspace{-2mm}
\end{table}

In Table~\ref{tab:metrics}, we present the semantic similarity analysis, comparing GLiNER and Qwen2.5-14B across key metrics. GLiNER outperformed Qwen2.5-14B in both Cosine Similarity (0.9808 vs. 0.9731) and Jaro-Winkler Similarity (0.8328 vs. 0.7601), demonstrating stronger semantic alignment and structural fidelity in its outputs. 

Qwen2.5-14B, however, achieved a lower Levenshtein Distance (0.4017 vs. 0.4358), indicating fewer character-level changes, which may benefit tasks requiring minimal textual modifications. 

Overall, GLiNER excels in preserving semantic and structural consistency, while Qwen2.5-14B offers advantages in minimizing textual alterations, highlighting their complementary strengths.

\section{Discussion}

The experimental findings confirm the effectiveness of the LegalGuardian framework in preserving privacy in legal applications of LLMs. By integrating NER models with structured masking and unmasking pipelines, the framework safeguards sensitive information while maintaining the utility of LLM-based tools in legal workflows.

GLiNER’s 100\% precision minimizes false positives, ensuring accurate redaction without over-masking—crucial in legal contexts where preserving non-sensitive information is important for document coherence. Conversely, Qwen2.5-14B's higher recall rate of 94\% demonstrates proficiency in detecting a broader range of entities, particularly in context-sensitive categories like addresses and personal names. This suggests Qwen's effectiveness in capturing overlooked entities, enhancing comprehensive redaction.

Entity-level accuracy metrics further highlight these distinctions. GLiNER achieved perfect precision and recall (100\%) for structured categories like tax IDs and case numbers, reflecting its strength in handling predictable patterns essential for compliance with legal standards. In contrast, Qwen2.5-14B outperformed GLiNER in recall for less structured categories, notably \textit{person} (77\% vs. 72\%) and \textit{address} (100\% vs. 98\%), indicating its advantage in contexts with linguistic variability.

Both models maintained strong semantic similarity between the unmasked outputs and the originals, with GLiNER slightly better preserving contextual integrity—critical where semantic fidelity is paramount. Minimal differences in semantic similarity indicate that masking and unmasking did not significantly distort the original content's meaning.

These findings imply that in high-risk legal scenarios, over-redaction—as facilitated by GLiNER's precision—may be preferable to reduce confidentiality breaches. Conversely, Qwen's robust recall demonstrates the utility of domain-specific LLMs in capturing a wider array of sensitive entities, albeit with a slight increase in false positives.

We recommend a hybrid approach within LegalGuardian, using an NER model like GLiNER as the primary tool to ensure high precision, supplemented by a local LLM like Qwen2.5-14B to enhance recall by capturing additional entities. This strategy balances precision and recall, providing comprehensive redaction without compromising semantic integrity or utility.

By integrating these complementary tools, LegalGuardian effectively addresses the privacy-utility trade-off in legal AI applications. It upholds attorney-client confidentiality while facilitating the responsible adoption of advanced AI technologies. As demonstrated by the comparative metrics (see Figure~\ref{fig:model-metrics}), the hybrid approach leverages the strengths of both models, proving that it's feasible to leverage LLM capabilities in legal contexts without compromising privacy. The hybrid method enhances sensitive information detection and masking efficacy while maintaining content quality and usefulness, essential for deploying AI technologies in legal workflows requiring strict confidentiality.

\section{Conclusion}

We addressed the challenge of preserving client confidentiality when integrating LLMs into legal practice by introducing LegalGuardian, a lightweight privacy-preserving framework for masking and unmasking PII in legal prompts. Leveraging NER techniques and local LLMs for PII extraction, our approach protects sensitive data before any external interaction, upholding the stringent confidentiality required in the legal profession.

Our experimental results show that LegalGuardian effectively balances privacy preservation with functional utilization of LLMs, indicating that legal professionals can harness advanced AI technologies without compromising document quality or ethical obligations.

LegalGuardian advances privacy-preserving technologies tailored for legal practices, offering a practical method for utilizing LLM capabilities while adhering to ethical responsibilities. By facilitating responsible AI integration, LegalGuardian paves the way for broader adoption and innovation in the legal industry.

Future work will focus on extending the framework to additional legal areas beyond immigration law, enhancing PII detection for more complex data. We plan to explore integration with cloud-based LLM services while maintaining strict privacy, potentially through techniques like secure multi-party computation or federated learning. We also aim to conduct user studies with practicing lawyers to assess LegalGuardian's practical usability and impact, refining the framework based on feedback.


\bibliography{paper}

\section{Appendix}

\begin{table}[!htb]
    \centering
    \begin{minipage}[!t]{0.48\linewidth} 
        \centering
        \begin{tabular}{|c|} 
            \hline
            \textbf{Document Titles} \\ 
            \hline
            Employer Support Letter \\
            Affidavit of Support \\
            Personal Statement \\
            Birth Certificate \\
            Marriage Certificate \\
            Tax Returns \\
            Work Authorization Document \\
            Medical Examination Report \\
            Police Clearance Certificate \\
            Immigration History Summary \\
            \hline
        \end{tabular} 
        \label{tab:document-titles} 
    \end{minipage}
    \hfill
    \begin{minipage}[!t]{0.48\linewidth} 
        \centering
        \begin{tabular}{|l|}
            \hline 
            \textbf{Task Types} \\ 
            \hline 
            Summarization \\
            Translation \\
            Legal Analysis \\
            Drafting \\
            \hline 
        \end{tabular} 
        \label{tab:task-types}
    \end{minipage}

    \caption{List of Document Titles and Task Types used in Synthetic Prompt Generation}
\end{table}

\begin{table}[!htb]
    \centering
    \begin{tabular}{|p{30mm}|p{50mm}|} 
        \hline 
        \textbf{Practice Area} & \textbf{Subfields} \\ 
        \hline 
        Visa Applications & Family-based visa, Employment-based visa, Student visa \\
        \hline 
        Green Cards & Adjustment of status, PERM processing \\
        \hline 
        Deportation Defense & Removal proceedings, Cancellation of removal \\
        \hline 
        Citizenship and Naturalization & Citizenship applications, Dual citizenship resolution \\
        \hline 
        Asylum and Refugee Law & Filing asylum applications, Defending refugees \\
        \hline 
        DACA & Initial applications, Renewals \\
        \hline 
        Employment Compliance & I-9 verification, E-Verify compliance \\
        \hline 
    \end{tabular} 
    \caption{Practice Areas with Subfields used in Synthetic Prompt Generation} 
    \label{tab:practice-areas}
\end{table}

\lstset{style=default_style}
\begin{listing*}[!htb]%
\caption{Fake PII generation using LLM}
\label{lst:random-generation-function1}%

\begin{lstlisting}[language=Python]
def generate_fake_text_with_llm(client, practice_area, subfield, document_title):
    prompt = f"""
    Generate a fake but realistic document excerpt for an immigration case based on the following scenario:

    Practice Area: {practice_area}
    Subfield: {subfield}
    Client Name: {entities['client_name']}
    Client Nationality: {entities['client_nationality']}
    Visa Type: {entities['visa_type']}
    Document Title: {entities['document_title']}

    Generate a paragraph of fake text only.  
    Provide the fake text strictly, without any explanations or additional content.
    """
    
    response = chat(model='qwen2.5:14b', messages=[{'role': 'user', 'content': prompt}])

    return response.message.content
\end{lstlisting}
\end{listing*}

\begin{listing*}[!htb]%
\caption{Fake prompt generation for various tasks using LLM}
\label{lst:random-generation-function2}%
\begin{lstlisting}[language=Python]
def generate_prompt(task_type, entities, subfield, practice_area, fake_text):
    if task_type == "Summarization":
        return { "prompt": f"My client, {entities['client_name']}, a {entities['client_nationality']} citizen holding {entities['visa_type']}, resides at {entities['home_address']}. This case, identified as {entities['case_id']}, involves their employer, {entities['employer_name']} (Tax ID: {entities['employer_tax_id']}), located at {entities['employer_address']}. Summarize the following document submitted as part of the case:\n\nDocument Title: {entities['document_title']}\n {fake_text}",
            "entities": entities }
    elif task_type == "Translation":
        return { "prompt": f"My client, {entities['client_name']}, a {entities['client_nationality']} citizen residing at {entities['home_address']}, has submitted a {entities['document_title']} written in English. Translate this document into Spanish to support their {subfield} case under {practice_area}:\n\n {fake_text}",
            "entities": entities }
    elif task_type == "Legal Analysis":
        return { "prompt": f"Analyze whether {entities['client_name']}, a {entities['client_nationality']} citizen residing at {entities['home_address']}, qualifies for {subfield} under {practice_area}. Consider the following details of their case:\n\n- Filing Date: {entities['filing_date']}\n- Case ID: {entities['case_id']}\n- Employer: {entities['employer_name']} (Address: {entities['employer_address']})\n- Key Facts: {fake_text}",
            "entities": entities }
    elif task_type == "Drafting":
        return { "prompt": f"Draft a {entities['document_title']} for my client, {entities['client_name']}, a {entities['client_nationality']} citizen holding {entities['visa_type']} and residing at {entities['home_address']}. This document supports their {subfield} case under {practice_area}. Include their employer details: {entities['employer_name']} (Tax ID: {entities['employer_tax_id']}, Address: {entities['employer_address']}).",
            "entities": entities }
\end{lstlisting}
\end{listing*}


\end{document}